\pdfoutput=1

\documentclass[11pt]{article}

\usepackage[]{acl}

\usepackage{times}
\usepackage{latexsym}

\usepackage[T1]{fontenc}

\usepackage[utf8]{inputenc}

\usepackage{microtype}

\usepackage{cellspace}
\newcommand\cincludegraphics[2][]{\raisebox{-\height}{\includegraphics[#1]{#2}}}
\usepackage{booktabs}
\usepackage{xcolor}
\definecolor{mylightgray}{HTML}{858585} 
\definecolor{second2}{HTML}{3700c8} 
\definecolor{second}{HTML}{4f00b0} 
\newlength\replength
\newcommand\ruleht{3pt}
\newcommand\repfrac{.3}
\newcommand\rulewidth{0.5pt}
\newcommand\drulefill{\leavevmode\dashfill\hfil%
\kern\dimexpr\repfrac\replength-\replength\relax}
\newcommand\dashfill[1][\repfrac]{\cleaders\hbox to \replength{%
\smash{\rule[\ruleht]{\repfrac\replength}{\rulewidth}}}\hfill}
\usepackage{microtype}
\usepackage{verbatim}
\usepackage{listings}

\usepackage{amssymb}
\usepackage{amsmath}
\usepackage{graphicx}

\usepackage{bm}
\newcommand{\vect}[1]{\bm{#1}}
\newcommand{\matr}[1]{\bm{#1}}

\newcommand{\vy}[0]{\vect{y}}

\newcommand{\vr}[0]{\vect{r}}

\newcommand{\vtheta}[0]{\vect{\theta}}

\newcommand{\mD}{\matr{D}}

\newcommand{\mY}{\matr{Y}}
\title{Improving the Factual Correctness of Radiology Report Generation with Semantic Rewards}

\author{\normalfont Jean-Benoit Delbrouck, Pierre Chambon, Christian Bluethgen, \\ Emily Tsai, Omar Almusa, Curtis P. Langlotz \\ Stanford University \\
  \texttt{jbdel@stanford.edu} }

\begin{document}
\maketitle
\begin{abstract}
Neural image-to-text radiology report generation systems offer the potential to improve radiology reporting by reducing the repetitive process of report drafting and identifying possible medical errors. These systems have achieved promising performance as measured by widely used NLG metrics such as BLEU and CIDEr. However, the current systems face important limitations. First, they present an increased complexity in architecture that offers only marginal improvements on NLG metrics. Secondly, these systems that achieve high performance on these metrics are not always factually complete or consistent due to both inadequate training and evaluation. Recent studies have shown the systems can be substantially improved by using new methods encouraging 1) the generation of domain entities consistent with the reference and 2) describing these entities in inferentially consistent ways. So far, these methods rely on weakly-supervised approaches (rule-based) and named entity recognition systems that are not specific to the chest X-ray domain. To overcome this limitation, we propose a new method, the RadGraph reward, to further improve the factual completeness and correctness of generated radiology reports. More precisely, we leverage the RadGraph dataset containing annotated chest X-ray reports with entities and relations between entities. On two open radiology report datasets, our system substantially improves the scores up to 14.2\% and 25.3\% on metrics evaluating the factual correctness and completeness of reports.
\end{abstract}

\section{Introduction}

An important medical application of natural language generation (NLG) is to build assistive systems that take X-ray images of a patient and generate a textual report describing clinical observations in the images~\cite{jing-etal-2018-automatic,NEURIPS2018_e0741335, chen2020generating, miura2021improving}. This is a clinically important task, offering the potential to reduce radiologists’ repetitive work and generally improve clinical communication~\cite{kahn2009toward}. \\

\begin{figure}[!t]
  \centering\includegraphics[width=1.0\linewidth]{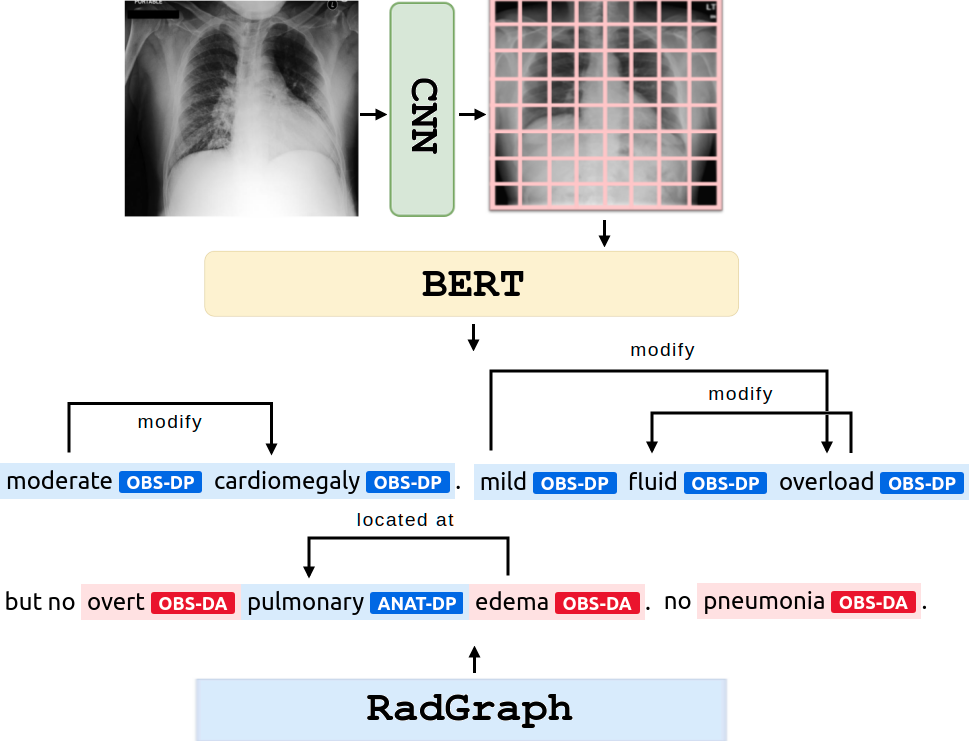}
  \caption{Overview of our radiology report generation pipeline. First, a neural network generates a radiology report given a chest X-ray image. We then leverage RadGraph to create semantic annotations of the output used to design reinforcement learning rewards.}
  \label{fig:radgraphxx}
\end{figure}

Recently, a lot of attention has been given to new architectures~\cite{chen2020generating, chen-etal-2021-cross-modal, ALFARGHALY2021100557} and how the structure of data could be input into the system~\citep{liu2021exploring}. These systems have achieved promising performance as measured by widely used NLG metrics such as BLEU~\cite{papineni2002bleu} and CIDEr~\cite{vedantam2015cider}. However, these studies face important limitations. First, they present an increased complexity in architecture that offers only marginal improvements on NLG metrics. Secondly, these systems that achieve high performance on NLG metrics are not always factually complete or consistent due to both inadequate training and evaluation of these systems. ~\citet{miura2021improving} have shown that existing systems are inadequate in factual completeness and consistency, and that an image-to-text radiology report generation (RRG) system can be substantially improved by replacing the widely used NLG metrics with "factually-oriented" methods encouraging 1) the generation of domain entities consistent with the reference and 2) describing these entities in inferentially consistent ways. So far, these new methods rely on weakly-supervised approaches (rule-based) to construct NLI models for radiology reports and  biomedical named entity recognition systems~\cite{zhang2021biomedical} that are not specific to chest X-rays. \\

Despite these "factually-oriented" methods being weakly supervised or being limited to generic biomedical entities, their use showed substantial improvements on a wide range of metrics and board-certified radiologists' evaluations. These findings motivate us to propose a new method to further improve the factual completeness and correctness of generated radiology reports. More precisely, we leverage RadGraph~\cite{8ffe9a5}, a dataset annotated by radiologists containing chest X-ray radiology reports along with annotated entities and relations. These annotations allow us to create two semantic graphs, one for the generated and one for the reference report. We then introduce three simple rewards that score the differences between the two graphs in terms of entities and relations. These rewards can be directly optimized using Reinforcement Learning (RL) to further improve the quality of the generated report by our systems. By doing so, we show on two popular chest X-ray datasets that our models are able to maximize the defined rewards but also outperform the previous works on various NLG and factually-oriented metrics. 

In summary our contributions are:
\begin{itemize}
    \item We propose a simple RRG architecture that 1) is fast to train and suitable for a RL setup and 2) performs equally well as the previous and more complex architectures proposed in the literature.
    \item We leverage the RadGraph dataset and the associated fine-tuned model to design semantic-based rewards that qualitatively evaluate the factual correctness and completeness of the generated reports.
    \item We show on two datasets that directly optimizing these rewards outperforms previous approaches that prioritize traditional NLG metrics.
\end{itemize}

The paper is structured as follows: first, we describe our factually-oriented graph-based rewards (\S \ref{sec:semantic_graph}). More precisely, we begin by examining the RadGraph dataset (\S~\ref{sec:radgraph}) and how we leveraged the annotations to create our rewards (\S~\ref{sec:rad_rew}). Then, we explain the architecture of the model (\S~\ref{sec:model}) that we used to generate reports and how we trained it using negative log-likelihood (NLL) and RL. The sections that follow afterwards are dedicated to the datasets used for the experiments (\S~\ref{sec:datasets}) and the metrics (\S~\ref{sec:metrics}) chosen to evaluate the generation of reports. This latter section is divided into two groups: the classic NLG metrics (\S~\ref{sec:metrics_nlg}) and the factually-oriented metrics (\S~\ref{sec:metrics_fact}). Finally, we present the results (\S~\ref{sec:results}) and end this paper with a section addressing related works (\S~\ref{sec:related}).

\section{Factually-oriented Graph-Based Reward} \label{sec:semantic_graph}

In this section, we present a new semantic graph-based reward, called the RadGraph reward, used throughout our experiments. We first start by explaining in Section \ref{sec:radgraph} the RadGraph dataset and how we get the annotations that shape our reward. In Section \ref{sec:rad_rew}, we explain how we construct the RadGraph reward and its different variants.

\subsection{RadGraph} \label{sec:radgraph}
\begin{figure}[!h]
  \centering\includegraphics[width=1.0\linewidth]{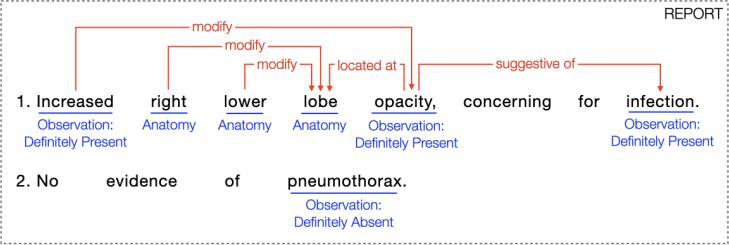}
  \caption{An example of a report annotated with entities and relations in the RadGraph dataset}
  \label{fig:ddd}
\end{figure}

RadGraph~\cite{8ffe9a5} is a dataset of entities and relations in full-text chest X-ray radiology reports based on a novel information extraction schema designed to structure radiology reports. The dataset contains board-certified radiologist annotations of 500 radiology reports from the MIMIC-CXR dataset ~\cite{2019arXiv190107042J}, which correspond in total to 14,579 entities and 10,889 relations. In addition, RadGraph also includes a test dataset of 100 radiology reports, split between two independent sets of board-certified radiologist annotations on reports from MIMIC-CXR and CheXpert~\cite{smit2020combining} datasets (50 reports each).

\paragraph{Entities} An entity is defined as a continuous span of text that can include one or more adjacent words. Entities in RadGraph center around two concepts: \textit{Anatomy} and \textit{Observation}. Three uncertainty levels exist for \textit{Observation}, leading to four different entities: \textit{Anatomy} (\textit{ANAT-DP}), \textit{Observation: Definitely Present} (\textit{OBS-DP}), \textit{Observation: Uncertain} (\textit{OBS-U}), and \textit{Observation: Definitely Absent} (\textit{OBS-DA}). \textit{Anatomy} refers to an anatomical body part that occurs in a radiology report, such as a ``lung''. \textit{Observation} refers to words associated with visual features, identifiable pathophysiologic processes, or diagnostic disease classifications. As an example, an \textit{Observation} could be “effusion” or more general phrases like “increased”. 

\paragraph{Relations} A relation is defined as a directed edge between two entities. Three levels exist: \textit{Suggestive Of (., .)}, \textit{Located At (., .)}, and \textit{Modify (., .)}. \textit{Suggestive Of (Observation, Observation)} is a relation between two \textit{Observation} entities indicating that the presence of the second \textit{Observation} is inferred from that of the first \textit{Observation}. \textit{Located At (Observation, Anatomy)} is a relation between an \textit{Observation} entity and an \textit{Anatomy} entity indicating that the \textit{Observation} is related to the \textit{Anatomy}. While \textit{Located At} often refers to location, it can also be used to describe other relations between an \textit{Observation} and an \textit{Anatomy}, such as shape or color. \textit{Modify (Observation, Observation)} or \textit{Modify (Anatomy, Anatomy)} is a relation between two \textit{Observation} entities or two \textit{Anatomy} entities indicating that the first entity modifies the scope of, or quantifies the degree of, the second entity. \\

The authors also released a PubMedBERT~\cite{gu2021domain} model fine-tuned on the RadGraph dataset. We leverage this trained model to create the annotations for the datasets used in our experiments. We will refer to this model as RadGraph model in what follows.

\subsection{RadGraph reward} \label{sec:rad_rew}

\begin{figure}[!h]
  \centering\includegraphics[width=1.0\linewidth]{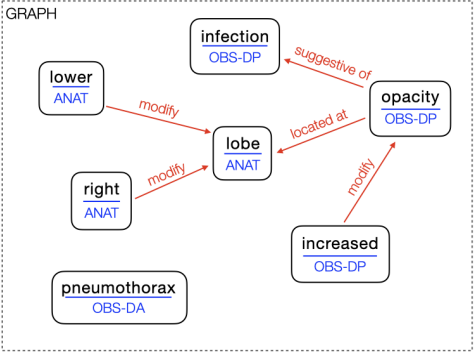}
  \caption{Graph view of the RadGraph annotations for the report in Figure~\ref{fig:ddd}.}
  \label{fig:radgraph}
\end{figure}

Using RadGraph annotation scheme and model, we design F-score style rewards that measure consistency and completeness of generated radiology reports compared to reference reports. Each of our rewards leverages the outputs of the released fine-tuned PubMedBERT model on RadGraph, namely the entities and the relations, on both a generated report and its reference.\\

The RadGraph annotations of a report can be represented as a graph $\mathcal{G}(V,E)$ with the set of nodes $V=\{v_1,v_2,\hdots, v_{|V|}\}$ containing the entities and the set of edges $E=\{e_1,e_2,\hdots, e_{|E|}\}$ the relations between pairs of entities. The graph is directed, meaning that the edge $e = (v_1, v_2) \ne (v_2, v_1)$. An example is depicted in Figure~\ref{fig:radgraph}. Each node or edge of the graph also has a label, which we denote as $v_{i_{L}}$ for an entity $i$ (for example "OBS-DP" or "ANAT") and $e_{ij_{L}}$ for a relation $e = (v_i, v_j)$ (such as "modified" or "located at"). We now proceed to describe three of our rewards.\\

\textbf{$\text{RG}_\text{E}$} \quad This reward focuses only on the nodes $V$. For the generated report $y$, we create a new set of node-label pairs $\bar{V}_y = \{{(v_i, v_{i_{L}})}\}_{i\in[1..{|V|}]}$ comprising all entities and their corresponding labels. We proceed to construct the same set for the reference report $\hat{y}$ and denote this set $\bar{V}_{\hat{y}}$. \\

\textbf{$\text{RG}_\text{ER}$} \quad This reward focuses on the nodes $V$ and whether or not a node has a relation in $E$. For the generated report $y$, we create a new set of triplets $\bar{V}_y = \{{(v_i, v_{i_{L}}, \epsilon_i)}\}_{i\in[1..{|V|}]}$. The value of $\epsilon_i$ is $1$ if $v_i$ has a relation in $E$ else 0. We proceed to construct the same set for the reference report $\hat{y}$ and denote this set $\bar{V}_{\hat{y}}$. \\

\textbf{$\text{RG}_{\overline{{\text{ER}}}}$ } \quad This reward focuses on the nodes $V$ and their relations in $E$. For the generated report $y$, we create a new set of tuples $\bar{V}_y = \{{(v_i, v_{i_{L}}, (v_i,v_j), e_{ij_{L}})} \mid i\in [1..|V|],\ j\in [1..|V|],\ j\ne i,\ (v_i, v_j) \in E\}$. In addition, for all the nodes $v_i$ with no relations, we include a tuple $(v_i, v_{i_{L}})$ in $\bar{V}_y$. We proceed to construct the same set for the reference report $\hat{y}$ and denote this set $\bar{V}_{\hat{y}}$.
\\ 

Finally, $\text{RG}_\text{E}$, $\text{RG}_\text{ER}$ and $\text{RG}_{\overline{{\text{ER}}}}$ are defined as the harmonic mean of precision and recall between their respective sets $\bar{V}_{\hat{y}}$ and $\bar{V}_{{y}}$. \\

As an illustration, we provide in Appendix~\ref{app:v_graph} the set $\bar{V}$ of the graph $\mathcal{G}$ in Figure~\ref{fig:radgraph}.

\section{Model} \label{sec:model}

\textbf{Architecture} \quad To encode an X-ray image, we extract convolutional visual features $\mD$ of size $49 \times 1024$ using a Densenet-121~\cite{huang2017densely}. To generate language, we use a one-layered BERT~\cite{vaswani2017attention,devlin-etal-2019-bert} with cross-attention over the visual features. More formally, the cross-attention of the transformer layer is written:
\begin{equation}
    \text{Cross-Attention}(Q, K, V) = \text{softmax}(\frac{QK^\top}{\sqrt{d}})V \label{eq:transfo_att}
\end{equation} where $Q$ is the BERT hidden state of size $d$ and $K$ and $V$ are the visual features $\mD$. The full detail of the model can be found at Appendix~\ref{app:model}.\\

\textbf{Training} \quad If we denote $\vtheta$ as the model parameters, then $\vtheta$ is learned by maximizing the likelihood of the observed report $\mY = (\vy_1, \vy_2, \cdots, \vy_n)$ or in other words by minimizing the negative log-likelihood. The objective function is given by:
\begin{equation}
\mathcal{L}(\vtheta) = - \sum\limits_{t=1}^n \log p_{\vtheta}(\vy_t | \vy<t, \mD) \label{eq:5}
\end{equation}

After the NLL training, we start a RL training that optimizes one of our factually-oriented rewards. The loss function in equation~\ref{eq:5} is now given by:

\begin{equation}
\mathcal{L}(\vtheta) = - \mathbb{E}_{\mY\sim p_{\vtheta}} \vr(\mY) 
\end{equation}

where $\vr(\mY)$ is the reward given to the generated report. We use the SCST algorithm~\cite{rennie2017self} to approximate the expected gradient of our non-differentiable reward function. The expression becomes:

\begin{equation}
\nabla_{\vtheta} \mathcal{L}(\vtheta) \approx - (\vr(\mY) - \vr(\bar{\mY})) \nabla_{\vtheta} \log_{p_{\vtheta}}(\mY)
\end{equation}

Here $\vr(\bar{\mY})$ acts as a baseline~\cite{sutton1998introduction} to reduce the variance of $\vr(\mY)$. In our case, $\vr(\bar{\mY})$ is the expected reward by sampling from the model during training.
\\

\textbf{Hyper-parameters} \quad Our model consists of 1 Transformer block of size 768 with a feed-forward layer of size 3072. As optimizer, we pick Adam~\cite{kingma2015adam} with a learning rate of $3e^{-4}$ and mini-batch size of 128. We decode with a beam-search of size 3.

\section{Datasets} \label{sec:datasets}
To carry out our experiments, we use two chest X-ray datasets: MIMIC-CXR~\cite{2019arXiv190107042J} and Open-i Chest X-ray (Indiana University, \citet{demner2012design}). For both datasets, we use the official splits but we discard the reports that do not contain a \textit{Findings} section. MIMIC-CXR thus consists of $152,173$ training samples, $1,196$ for validation and $2,347$ for testing; similarly, the Indiana dataset, originally containing $5,935$ training images, $740$ for validation and $740$ for testing, consists of only $3,335$ reports in total if we do not count the multiplicity of images in each study and if we discard the reports without \textit{Findings}. Since this renders the dataset too small to both train and test a RRG model, Open-i dataset is only used for testing purposes.\\

For each sample of both datasets, we generate what we consider the ground-truth radiology diagnostic by running the CheXbert labeler~\cite{smit2020combining} on the ground-truth \textit{Findings} section. This creates for each ground-truth report the associated diagnostic label, which describe for 14 possible observations the degree of presence (e.g. consolidation or edema for which the report can be positive, negative, uncertain or unspecified). This label is used to compute the F1 CheXbert score (see Section \ref{sec:metrics_fact}).

\section{Metrics} \label{sec:metrics}
In this section, we proceed to report all the metrics used to evaluate the \textit{Findings} generated by our model against the human-redacted \textit{Findings}. We divided our metrics into two categories, the NLG metrics as widely reported in the NLG literature, and the factually-oriented metrics, specific to the evaluation of factual correctness and completeness of radiology reports. \\

\subsection{NLG-oriented} \label{sec:metrics_nlg}

We report the BLEU~\cite{papineni2002bleu} and CIDEr~\cite{vedantam2015cider} metrics to evaluate the generations. To be consistent with previous work~\cite{chen-etal-2021-cross-modal}, we also report the ROUGE-L metric~\cite{lin2004rouge}.

\subsection{Factually-oriented}\label{sec:metrics_fact}
The following presented metrics evaluate the factual correctness and completeness of the generated \textit{Findings} in different ways. We proceed to describe them and their differences. \\

\textbf{$\text{fact}_\text{ENT}$}~\cite{miura2021improving} \quad A named entity recognizer (NER) is applied to the generated report $\hat{y}$ and the corresponding reference $y$, giving respectively two sets of extracted entities $\mathbb{E}_{\hat{y}}$ and $\mathbb{E}_{{y}}$. $\text{fact}_\text{ENT}$ is defined as the harmonic mean of precision and recall between the two sets $\mathbb{E}_{\hat{y}}$ and $\mathbb{E}_{{y}}$. The clinical model of Stanza~\cite{qi2020stanza} is used as NER. \\

\begin{table*}[!t]
	\centering
	\begin{tabular}{lcccccccc}
		\multicolumn{1}{c}{\bf Model}  &\multicolumn{6}{c}{\bf 				Test Scores}
		\\ \hline \\
		&BL4 & ROUGEL & $\text{F}_1$cXb & $\text{fact}_\text{ENT}$ & $\text{fact}_\text{ENTNLI}$ &  $\text{RG}_\text{E}$ &  $\text{RG}_\text{ER}$ &  $\text{RG}_{\overline{{\text{ER}}}}$ \\
		\textbf{MIMIC-CXR} \\
		\multicolumn{9}{c}{{\color{mylightgray}\drulefill \quad \textit{Using NLL} \quad \drulefill}} \\
		\citeauthor{pmlr-v106-liu19a} \shortcite{pmlr-v106-liu19a}  & 7.6 & --- & 29.2 & --- & --- & --- & --- & ---  \\
		\citeauthor{chen2020generating} \shortcite{chen2020generating}  & 8.6 & 27.7 & 34.6 & --- & --- & --- & --- & ---  \\
		\citeauthor{chen-etal-2021-cross-modal} \shortcite{chen-etal-2021-cross-modal}  & 10.6 & \textbf{27.8} & 40.5 & --- & --- & --- & --- & ---  \\
		\citeauthor{miura2021improving} \shortcite{miura2021improving} & \textbf{11.5} & ---  & 44.7 &  27.3 & 24.4 & --- & --- & --- \\
		ours (NLL) & 10.5 & 25.3  & \textbf{44.8} &  \textbf{28.0} & \textbf{26.8} & 23.0 & 20.2 & 15.3 \\
		\multicolumn{9}{c}{{\color{mylightgray}\drulefill \quad \textit{Using RL} \quad \drulefill}} \\
		\citeauthor{miura2021improving} ($\text{fact}_\text{ENT}$) & 11.1 & --- &{ 56.7}  & 39.5 &  34.8 & --- & --- & --- \\
		\citeauthor{miura2021improving} ($\text{fact}_\text{ENTNLI}$) & 11.4 & --- &  {56.7}  &  38.5 &  37.9 & --- & --- & ---\\
		ours ($\text{RG}_\text{E}$) & 11.4 & 26.3 & 59.4 & 41.2 & 42.7 & {36.8} & 32.5 & 21.5 \\
		ours ($\text{RG}_\text{ER}$) & {11.4} & \textbf{26.5}  & \textbf{62.2} &  \textbf{42.5} & \textbf{43.3} & \textbf{37.1} & \textbf{34.7} & 23.5 \\
		ours ($\text{RG}_{\overline{{\text{ER}}}}$) & \textbf{11.6} & 25.9  & 51.4 &  41.8 & 40.9 & 35.4 & 31.6 & \textbf{23.8} \\
		\hline \\
		\textbf{Open-i} \\
		\multicolumn{9}{c}{{\color{mylightgray}\drulefill \quad \textit{Using NLL} \quad \drulefill}} \\
		\citeauthor{chen-etal-2021-cross-modal} \shortcite{chen-etal-2021-cross-modal} & 12.0 &\textbf{ 29.8} & --- &  --- &  --- &  --- &  --- &  ---\\
		\citeauthor{miura2021improving} \shortcite{miura2021improving} & \textbf{12.1} & 28.8  & 32.2 &   \textbf{40.6}&  \textbf{42.9} &  --- &  --- &  ---\\
		ours (NLL) & 11.4 & ---  & \textbf{33.1} &   \textbf{40.6}&  42.6 &  29.2 &  26.4 &  18.1 \\
		\multicolumn{9}{c}{{\color{mylightgray}\drulefill \quad \textit{Using RL} \quad \drulefill}} \\
		\citeauthor{miura2021improving} ($\text{fact}_\text{ENT}$) & 12.0 & --- & 48.3  & 44.4 &  46.8 & --- & --- & ---  \\
		\citeauthor{miura2021improving} ($\text{fact}_\text{ENTNLI}$) & 13.1 & --- & {47.8}  &  43.6&  47.1 & --- & --- & ---\\
		ours ($\text{RG}_\text{E}$) & 13.1 & 32.5 & 46.8 & 44.3 & 51.2 & \textbf{44.1} & 38.8 & 29.9 \\
		ours ($\text{RG}_\text{ER}$) & \textbf{13.9} & \textbf{32.7} & \textbf{49.1} & \textbf{46.0} & \textbf{58.9} & 43.6 & \textbf{41.2} & 31.9 \\
		ours ($\text{RG}_{\overline{{\text{ER}}}}$) & 12.1 & 30.6 & 45.3 & 43.3 & 53.2 & 41.9 & 39.1 & \textbf{32.2}
		
	\end{tabular}      
	\caption{Comparison of our models against the state of the art (we took the best four models performing on MIMIC-CXR and the two best on Open-i). Results reported for Open-i are from models trained on MIMIC-CXR and tested on the entirety of the Open-i dataset. BL4 and $\text{F}_1$cXb refers to the BLEU4 and $\text{F}_1$CheXbert metrics. In \textbf{bold} are highlighted the best scores per category (i.e. NLL/RL and MIMIC-CXR/Open-i) }
    \label{score-tabular}
\end{table*}%

\textbf{$\text{fact}_\text{ENTNLI}$}~\cite{miura2021improving} \quad  This score is an extension of \textbf{$\text{fact}_\text{ENT}$} with Natural Language Inference (NLI). Here, an entity $e$ of $\mathbb{E}_{\hat{y}}$ is not automatically considered correct if present in $\mathbb{E}_{{y}}$. To be considered valid, the sentence $s_{\hat{y}}$ containing entity $e$ must not present a contradiction with its counterpart sentence $s_{{y}}$ in the reference report. The counterpart sentence $s_{{y}}$ in the reference report is the sentence with the highest BERTScore~\cite{zhang2019bertscore} against $s_{\hat{y}}$. The NLI model outputs whether sentence $s_{{y}}$ is a contradiction of $s_{\hat{y}}$. We use the NLI model weights of \citet{miura2021improving}, which relies on a BERT-architecture. \\

\textbf{F$_1$CheXbert}~\cite{zhang2020optimizing} \quad This score uses CheXbert~\cite{smit2020combining}, a Transformer-based model trained to output abnormalities (fourteen classes) of chest X-rays given a radiology report as input. F$_1$CheXbert is the F1-score between the prediction of CheXbert over the generated report $\hat{y}$ and the corresponding reference $y$. To be consistent with previous works, the score is calculated over 5 observations: atelectasis, cardiomegaly, consolidation, edema and pleural effusion. \\
	
\textbf{RG rewards} \quad We use the rewards explained in Section \ref{sec:rad_rew} as evaluation scores.

\subsection{RadGraph vs $\text{fact}_\text{ENT}$ and $\text{fact}_\text{ENTNLI}$}

Theoretically, $\text{RG}_\text{E}$ encapsulates both $\text{fact}_\text{ENT}$ and $\text{fact}_\text{ENTNLI}$ concepts. Indeed, $\text{RG}_\text{E}$ focuses on having the right entities and also the right entity labels. Given that the label of an entity contains the notion of an anatomy or observation, the former being always present by definition and the latter having a degree of presence ("present", "absent" or "uncertain"), $\text{RG}_\text{E}$ can penalize a report presenting a contradiction with the reference, in the same fashion $\text{fact}_\text{ENTNLI}$ does. \\

Moreover, $\text{RG}_\text{E}$ presents two more advantages. First, it does not rely on an external model, such as the BERTScore, to map a hypothesis sentence with its counterpart in the reference report to run the NLI model. Secondly, the entities and entity labels evaluated by $\text{RG}_\text{E}$ are computed by a NER model specifically trained on chest X-rays, while $\text{fact}_\text{ENT}$ relies on a general-purpose biomedical NER system.

\section{Results} ~\label{sec:results}

In this section, we discuss the results displayed in Table~\ref{score-tabular}. We divide the section into Quantitative analysis (Section \ref{sec:quant}), Qualitative analysis (Section \ref{sec:qual}) and Limitations (Section \ref{sec:limit}).

\subsection{Quantitative analysis} \label{sec:quant}

\textbf{Using NLL} \quad The models reported in the NLL section are trained using only the negative log likelihood loss. We can see that our simple approach, referred to as "ours (NLL)", performs similarly on the NLG metrics (better in BLEU but worse in ROUGEL compared to ~\citet{chen-etal-2021-cross-modal}), but outperforms previous works on the factually-oriented metrics. On MIMIC-CXR, our baseline is up 2.3\% on  $\text{fact}_\text{ENT}$ and up 8.0\% on $\text{fact}_\text{ENTNLI}$ compared to~\citet{miura2021improving}. \\

\begin{table*}[!t]
\resizebox{\textwidth}{!}{
\begin{tabular}{Sc Sc Sc Sc}
\hline
Image & Ours (NLL) & Ours ($\text{RG}_\text{ER}$) & Reference \\ \hline
\cincludegraphics[width=0.18\linewidth]{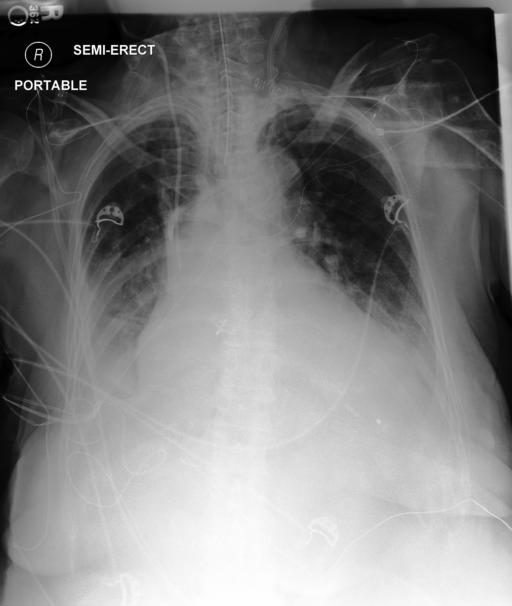} & \cincludegraphics[width=0.25\linewidth]{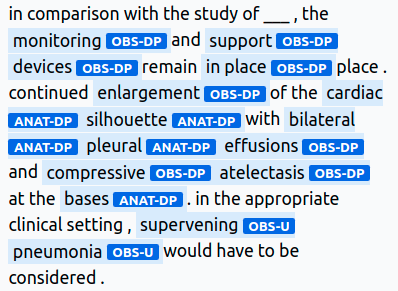} &  \cincludegraphics[width=0.25\linewidth]{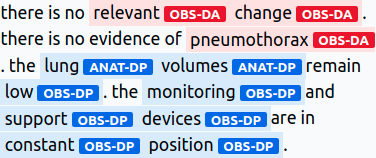} &   
\cincludegraphics[width=0.25\linewidth]{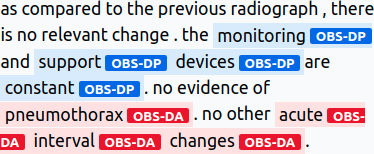} \\ \hline
       $\text{RG}_\text{ER}$ reward: &   26.0\%       &  52.6\%      &     \\ \hline
      ChexBert labels        &  \small \begin{tabular}[c]{@{}l@{}} Cardiomegaly, Pneumonia, \\ Atelectasis Support Devices  \end{tabular} &  \small Support Devices, No Finding     &   \small Support Devices, No Finding  \\ \hline 
    \cincludegraphics[width=0.20\linewidth]{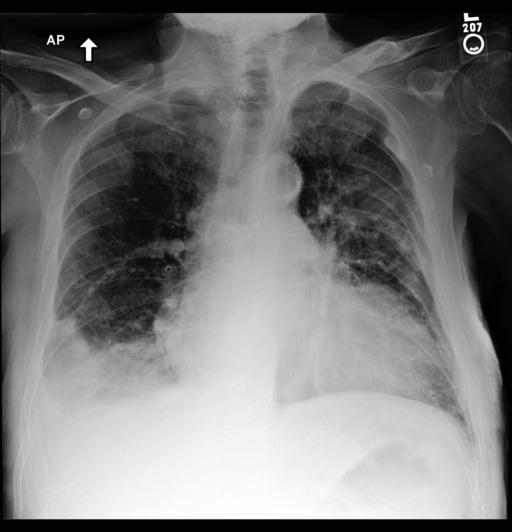} & \cincludegraphics[width=0.25\linewidth]{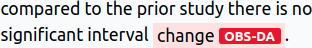} &  \cincludegraphics[width=0.25\linewidth]{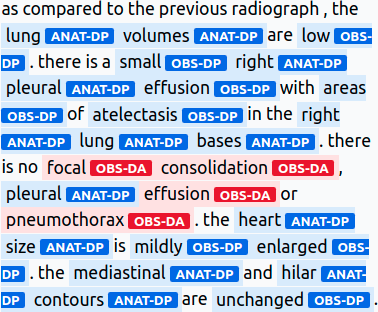} &   
\cincludegraphics[width=0.25\linewidth]{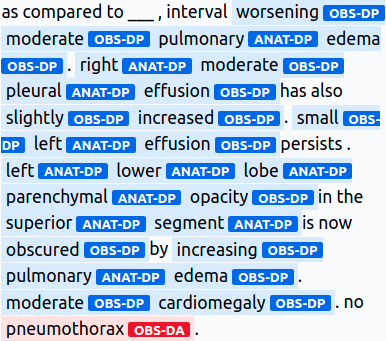} \\ \hline
       $\text{RG}_\text{ER}$ reward: &   0.0\%       &  22.2 \%      &     \\ \hline
      ChexBert labels &  \small No findings &  \small  \begin{tabular}[c]{@{}l@{}} Enlarged Cardiomediastinum, \\ Cardiomegaly \end{tabular}     &   \small \begin{tabular}[c]{@{}l@{}} Enlarged Cardiomediastinum, \\ Cardiomegaly Lung Opacity, \\ Pleural Effusion\end{tabular}  \\ \hline 
\end{tabular}
}
\caption{Cherry picked examples that compare two of our models' outputs: Ours (NLL) and Ours ($\text{RG}_\text{ER}$).}
\label{table:cherry_picked}
\end{table*}

\textbf{Using RL} \quad These results are the main contribution of our paper. We notice that optimizing any of the presented rewards in Section~\ref{sec:rad_rew} improved the RadGraph scores, validating the design of our rewards. Our best performing model is RG$_\text{ER}$ accross both datasets. On MIMIC-CXR,  it shows an improvement of respectively 61.3\%, 71.7\% and 53.5\% on the RG$_\text{E}$, RG$_\text{ER}$ and $\text{RG}_{\overline{{\text{ER}}}}$ metrics compared to our NLL baseline. This model also reports the best scores on the NLG metrics as well as improvements of +7.59\% on $\text{fact}_\text{ENT}$ and +14.2\% on $\text{fact}_\text{ENTNLI}$ over \citet{miura2021improving}. This means reports are generated with more factually-correct entities and less contradictions. On the {F$_1$CheXbert} score, our model also reports an improvement of +9.7\% compared to the $\text{fact}_\text{ENTNLI}$ model. It is also worth noting that if we were to include all abnormalities in the computation of the F$_1$CheXbert, $\text{RG}_{{{\text{ER}}}}$ score would be \textbf{56.0\%}, meaning that our model also performs well on less represented classes. \\

On the out-of-domain test-set of open-i, our model trained with RG$_\text{ER}$ reward reports a similar trend, with a noticeable improvement of 16.7\% on the $\text{fact}_\text{ENTNLI}$ metric (55.0 vs 47.1). \\

\textbf{RG$_\text{ER}$ vs $\text{RG}_{\overline{{\text{ER}}}}$} \quad Surprisingly, RG$_\text{ER}$ outperforms $\text{RG}_{\overline{{\text{ER}}}}$ on both datasets and on every metrics. A hypothesis is that the  $\text{RG}_{\overline{{\text{ER}}}}$ reward is too restrictive on the relations. Indeed, for an entity and its relation, a point of precision is only given if the label of the relation and the target entity of the relation are both found in the reference report. On MIMIC-CXR test-set, only 20.4\% of the relations generated by our model are correct (5071 out of 24818 generated relations). It means that for 79.6\% of the generated relations, our model received a negative signal even if some of these relations were partly correct. By contrast, when optimizing  RG$_\text{ER}$ where a point of precision is given when the relation for an entity exists in the generated and reference report, regardless of the label and the target entity, 38.7\% of the relations are correct (9974 out of 25769 generated relations). This is 18.4\% more than $\text{RG}_{\overline{{\text{ER}}}}$. We assume that this relaxed constraint encourages the model to generate relations and therefore more factually correct and complete reports.


\textbf{{Impression} section} \quad We also evaluate our models on their potential to generate the \textit{Impression} section of a report, based on the corresponding chest X-ray image, instead of the \textit{Findings} section (see Appendix~\ref{app:impression}). \textit{Impression} highlights the key observations and conclusions of the radiology study. Automating this task is also critical because the \textit{Impression} section is the most important part of a radiology report, and can be time-consuming and error-prone to produce. In addition, generating \textit{Impression} is related to Radiology Report Summarization~\cite{zhang2020optimizing} where a system has to summarize the \textit{Findings} section of a report into \textit{Impression}, making this choice even more relevant. MIMIC-CXR now consists of $185,816$ and $1,521$ samples for the training and validation sets. The MIMIC-CXR and Open-i test sets now have respectively $2,224$ and $3,820$ samples.

\subsection{Qualitative analysis} \label{sec:qual}
First, we performed a human evaluation to further confirm whether the generated radiology reports are more
factually complete and consistent. Two board-certified radiologists were asked to perform up to a hundred studies, where they had to choose between two findings given the chest X-ray. The two findings are from $\text{fact}_\text{ENTNLI}$ and our model $\text{RG}_\text{ER}$. On average, the radiologists favored our model. We give more details of the study in Appendix~\ref{app:study}.
\\

The Figure \ref{fig:numbers_of_elements} shows the number of entities and relations generated by our best model $\text{RG}_\text{ER}$, aggregated per label, on the MIMIC-CXR test-set. The takeaways are that 1) our model generates 20\% more \textit{Anatomy} entities compared to the ground-truth reports 2) for the 4 most frequent labels, namely \textit{OBS-DP}, \textit{modify}, \textit{located\_at}, \textit{OBS-DA}, the model generates between -12\% and +24\% entities and relations 3) our model barely generated any occurrences of the two most under-represented labels: \textit{OBS-U} and \textit{suggestive\_of}. \\

It is also interesting to note that the median word-length of the $\text{RG}_\text{ER}$ findings is 7\% lower than the ground-truth findings and 25\% lower for the NLL findings (on MIMIC-CXR). We can see in Table \ref{table:cherry_picked} two cherry picked examples showing reports generated by our two models. We see in both instances that the length of the reports from our $\text{RG}_\text{ER}$ model is closer to the references lengths compared to the NLL model.






\begin{table*}[!t]
\setlength{\tabcolsep}{4.7pt}
\renewcommand{\arraystretch}{1.2}
	\centering
	\begin{tabular}{lcccccccccccccc}
	        \hline
	    \multicolumn{15}{c}{\textbf{MIMIC-CXR}}\\
        \hline
    & \multicolumn{2}{c}{\hspace{-0.3em}ANAT-DP} & \multicolumn{2}{c}{\hspace{-0.3em}OBS-DP} & \multicolumn{2}{c}{\hspace{-0.3em}modify} & \multicolumn{2}{c}{\hspace{-0.3em}located\_at} & \multicolumn{2}{c}{\hspace{-0.3em}OBS-DA} & \multicolumn{2}{c}{\hspace{-0.3em}OBS-U} & \multicolumn{2}{c}{\hspace{-0.3em}suggestive\_of}\\
    & Prec. & 
    \hspace{-0.6em}Rec. & Prec. & \hspace{-0.6em}Rec. & Prec. & \hspace{-0.6em}Rec. & Prec. & \hspace{-0.6em}Rec. & Prec. & \hspace{-0.6em}Rec. & Prec. & \hspace{-0.6em}Rec. & Prec. & \hspace{-0.6em}Rec. \\
        \hline
    ours (NLL) 
    & 29.7 & \hspace{-0.6em}23.6 
    & 26.6 & \hspace{-0.6em}15.7 
    & 13.6 & \hspace{-0.6em}9.2 
    & 14.2 & \hspace{-0.6em}11.1 
    & 29.8 & \hspace{-0.6em}48.4
    & 13.5 & \hspace{-0.6em}3.0 
    & 9.8 &  \hspace{-0.6em}2.1\\
            \hline
    ours ($\text{RG}_\text{ER}$) 
    & 38.0 & \hspace{-0.6em}45.5 
    & 33.6 & \hspace{-0.6em}29.8 
    & 16.5 & \hspace{-0.6em}14.6 
    & 18.3 & \hspace{-0.6em}22.8 
    & 49.5 & \hspace{-0.6em}47.3 
    & 0.0 &  \hspace{-0.6em}0.0 
    & 11.1 & \hspace{-0.6em}0.1\\
	\end{tabular}
\caption{\centering Precision and recall per each entity and relation label on the MIMIC test-set, for the $\text{RG}_\text{ER}$ model. For each label and study, we assess whether the true words from the ground-truth report correspond to the generated words from the generated report - or pairs of source and target words in the case of relations.}
\label{table:recall_precision_entities_relations}
\end{table*}

\section{Related work} ~\label{sec:related}

First, we describe the studies that presented architectural novelties. Usually, they focus on improving the widely used NLG metrics such as BLEU and ROUGE. We then proceed to go over the previous works that used Reinforcement Learning (RL) to optimize NLG or factually-oriented metrics. Finally, we quickly mention a few projects that do not fall into the first two categories. 
\\

\textbf{Architectural novelties} \quad \citet{chen2020generating} proposed to generate radiology reports with memory-driven Transformer, where a relational memory is designed to record key information of the generation process and a memory-driven conditional layer normalization is applied to incorporating the memory into the decoder of Transformer. In ~\citet{chen-etal-2021-cross-modal}, authors investigated cross-modal memory networks to enhance the encoder-decoder framework for radiology report generation, where a shared memory is designed to record the alignment between images and texts so as to facilitate the interaction and generation across modalities. \citet{liu2021exploring} used Posterior-and-Prior knowledge to imitate the working patterns of radiologists, who first examine the abnormal regions and assign the disease topic tags to the abnormal regions, and then rely on the years of prior medical knowledge and prior working experience accumulations to write reports. More specifically, the prior knowledge consists of a medical knowledge graph built using unsupervised topic modeling. Topics are defined as nodes and are grouped by the organ or body part that they relate to. They connect their nodes with bidirectional edges, resulting in closely connected related topics. Another notable work used a Knowledge Graph Auto-Encoder~\cite{liu2021auto} taking as input a knowledge graph constructed in an unsupervised manner. \\

We differ from these two last works in two ways: 1) our semantic graph is generated by a model trained on high-quality human-annotated data and 2) these studies used the graph as input to their model while we use our graph as an evaluation metric that can be directly optimized using Reinforcement Learning. \\

\textbf{Reinforcement Learning (RL)} \quad \cite{pmlr-v106-liu19a} improved Radiology Report Generation by optimizing the correctness of the output of CheXpert. The metric they optimized is equivalent to the CheXbert metric presented in our paper. ~\cite{miura2021improving} proposed new metrics to evaluate the factual correctness and consistency of the generated report. The metrics are included in our work and referred to as 
$\text{fact}_\text{ENT}$ and $\text{fact}_\text{ENTNLI}$ in Section~\ref{sec:metrics_fact}. \\

Our paper is an original contribution to these two preceding works: we present a new evaluation metric that answers previous weaknesses. First, our graph-based annotations are generated by a model trained on a dataset with entities and relations labeled by radiologists. Second, our annotations are specific to the chest X-ray domain. Finally, our reward captures new semantic nuances such as new entity labels (anatomy, observation, absent, present) and new relationship levels between words. \\

\textbf{Other works} \quad Two previous works generated radiology reports using retrieval methods. ~\citet{endo2021retrieval} used a retrieval-based radiology report generation approach using a
pre-trained contrastive language-image model. At test time, they retrieved the most likely report in the training dataset given the representation of the encoded X-ray. ~\citet{NEURIPS2018_e0741335} employed a hierarchical decision-making procedure. For each sentence, a high-level retrieval policy module chooses to either retrieve a template sentence from an off-the-shelf template database, or invoke a low-level generation module to generate a new sentence. The decisions are updated via reinforcement learning, guided by sentence-level and word-level rewards. Finally, ~\citet{yang2021knowledge} decided to input the RadGraph annotations in addition to the X-ray image and showed moderate to no improvement compared to previous works.

\section{Conclusion}

In this paper, we leveraged the RadGraph dataset containing annotated chest X-ray reports with entities and relations between entities to design a new reward that qualitatively evaluate the factual correctness and completeness of the generated reports. We showed on two datasets that directly optimizing these rewards outperforms previous approaches that prioritize traditional NLG metrics or leverage unsupervised out-of-domain systems as factual-oriented metrics. Our best model reports up to +14.2\% improvements on these factual metrics on the MIMIC-CXR and +25.3\% on Open-i.

\section{Limitations} \label{sec:limit}

In this section, we highlight four limitations of our work. \\

First, rewards based solely on entities~\cite{miura2021improving} or entities and relations cannot be optimized without counter-effects happening. Indeed, optimizing $\text{fact}_\text{ENT}$ or RG$_\text{ER}$ will encourage the model to discard the grammar and generate reports such as "left lower right base uper opacities pleural cardiopulmonary cardiopulmonary atelectasis" to maximize the precision of entities generated. For this reason, we follow the settings of previous work and optimize one of our RG metrics alongside the BERTScore~\cite{zhang2019bertscore} and the NLL loss (with weights 0.495, 0.495 and 0.01 respectively)  \\

\begin{figure}[!h]
  \centering\includegraphics[width=1.0\linewidth]{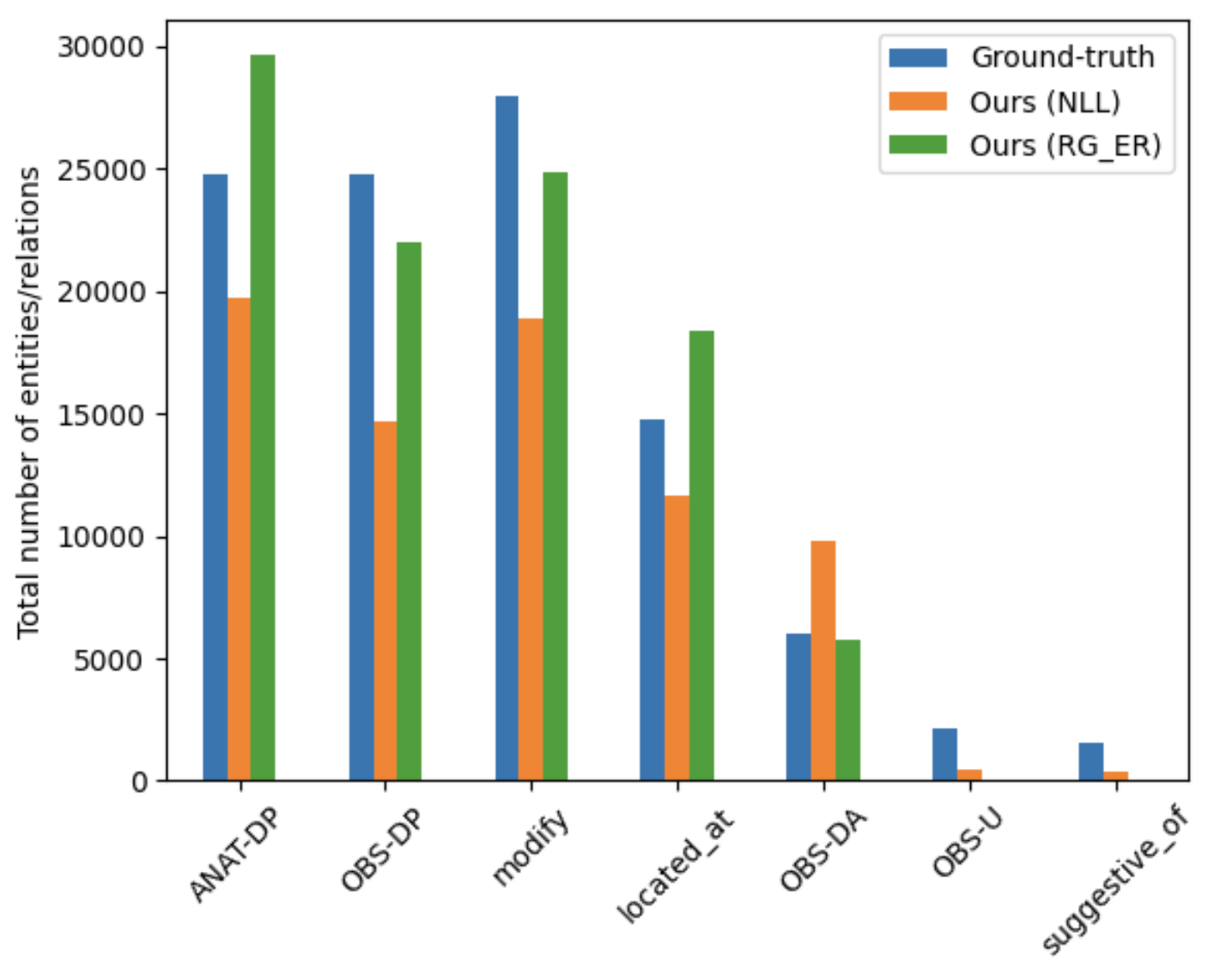}
  \caption{Number of entities and relations generated by our best model $\text{RG}_\text{ER}$, aggregated per label, on the MIMIC-CXR test-set. We also show the ground-truth distribution by running the RadGraph model on the reference test-set.}
  \label{fig:numbers_of_elements}
\end{figure}

Secondly, our model capability at correctly connecting observations and corresponding anatomies remains limited. Table \ref{table:recall_precision_entities_relations} depicts the precision and recall per entity and relation labels: in general, we observe that precision and recall have similar values among each label, but vary significantly from one label to the other. Excluding the two under-represented labels, \textit{OBS-U} and \textit{suggestive\_of}, entities have a macro-averaged recall of 40.9\% compared to only 18.7\% for relations.  \\


More critically, the \textit{OBS-U} entities are not correctly learned by our $\text{RG}_\text{ER}$ model, as underlined in Table~\ref{table:recall_precision_entities_relations}, the recall for this label being 0. We measured that 30\% of the words labeled as \textit{OBS-U} in the reference are incorrectly generated as \textit{OBS-DP} or \textit{OBS-DA} by our mode. The rest of the "missed" \textit{OBS} and \textit{ANAT} entities are due to the entity being not present in the generated report. Concerning the errors on the relation labels, we noticed that 15\% of the relations have the incorrect relation label, while the rest of the errors are due to the relation being just absent. \\

Finally, we note that even though our model fits on a single GPU of 12 GB, training the model using RL is computationally expensive. A RL epoch is between 7 to 10 hours on MIMIC-CXR (depending on the randomness of the sampling) against 50 minutes for NLL training.

\clearpage

\bibliography{custom}

\appendix

\clearpage

\section{Code release} ~\label{sec:discussion}
To help with further research, we make our code publicly available using the ViLMedic library~\cite{delbrouck-etal-2022-vilmedic}. More specifically, we release the code of all the factually-oriented metrics presented in Section~\ref{sec:metrics_fact} in one package. Our code also includes SCST~\cite{rennie2017self} on top of the widely-used NLP library HuggingFace~\cite{wolf2020transformers}. We hope this effort will improve reproducibility of the factually-oriented metrics and allow for a fairer comparison of the performance of future radiology report generation systems. Our code is available at \url{https://github.com/jbdel/vilmedic}.

\section{Set $\bar{V}$ of Figure~\ref{fig:radgraph}}\label{app:v_graph}

$V=\{$lower, infection, right, lobe, opacity, pneumothorax, increased$\}$\\
$E=\{$(right, lobe), (lower, lobe), (opacity, infection), (opacity, lobe), (increased, opacity)$\}$\\

$\bar{V}$ of RG$_{\text{E}}=\{$(lower, anat), (infection, obs-dp), (right, anat), (lobe, anat), (opacity, obs-dp), (increased, obs-dp), (pneumothorax, obs-da)$\}$\\

$\bar{V}$ of RG$_{\text{ER}}=\{$(lower, anat, 1), (infection, obs-dp, 0), (right, anat, 1), (lobe, anat, 0), (opacity, obs-dp, 1), (increased, obs-dp, 1), (pneumothorax, obs-da, 0)$\}$\\

$\bar{V}$ of RG$_{\bar{\text{ER}}}=\{$(lower, anat, lobe, modify), (infection, obs-dp), (right, anat, lobe, modify), (lobe, anat), (opacity, obs-dp, infection, suggestive of), (opacity, obs-dp, lobe, located\_at), (increased, obs-dp, opacity, modify), (pneumothorax, obs-da)$\}$\\

\section{Qualitative Study} \label{app:study}

To evaluate qualitatively how our $\text{RG}_\text{ER}$ model compares to previous models such as $\text{fact}_\text{ENTNLI}$ model, we built a study that asked radiologists to assess for each test image, based on their experience and the clinical expectations, which corresponding generated report is the best.\\

The clinical studies chosen for this experience are from the MIMIC-CXR test set and contain the following labels:

\begin{lstlisting}[
    basicstyle=\small,
    caption={Labels of the 100 clinical studies selected for the qualitative comparison of RRG models. There is more than 100 labels since one report can contain multiple labels. We made sure to have at least 10 instances for each abnormality.},captionpos=b,
     label={listing:solution}
]
{
    "Lung Opacity": 39,
    "Pleural Effusion": 36,
    "Support Devices": 36,
    "Atelectasis": 26,
    "Cardiomegaly": 19,
    "Lung Lesion": 19,
    "Pleural Other": 16,
    "Enlarged Cardiomediastinum": 15,
    "Pneumonia": 15,
    "Fracture": 14,
    "Edema": 13,
    "Consolidation": 13,
    "Pneumothorax": 10,
    "No Finding": 5
}
\end{lstlisting}

The radiologists received the following instructions: "Please evaluate the radiology reports given the chest x-ray using the following criteria: 1. factual correctness (is it correct ?); 2. factual completeness (how complete is the report ?); 3. factual consistency (is there any contradiction within the report ?)".

Radiologists were asked to choose, for each pair of generated reports, one being from $\text{fact}_\text{ENTNLI}$ model and the other from $\text{RG}_\text{ER}$ model, a score on a scale ranging from -2 (report 1 is much better) to 2 (report 2 is much better). Then, we aggregated the scores for each pair of reports and for each labeler, and computed the average score of reports coming from $\text{RG}_\text{ER}$ model compared to reports coming from $\text{fact}_\text{ENTNLI}$ model.  \\

Compared to $\text{fact}_\text{ENTNLI}$, our $\text{RG}_\text{ER}$ model is: \\

\small much worse \quad slightly worse \quad slightly better \quad much better\\
\includegraphics[width=1.0\linewidth]{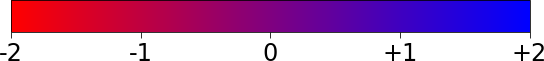} \\

{\centering Reader 1: \colorbox{second}{\textcolor{white}{0.485 $\pm$ 0.662}}  \quad Reader 2: \colorbox{second2}{\textcolor{white}{0.891 $\pm$ 0.461}} } 

\vspace{1cm}
\normalsize

According to both radiologists, on average reports from our $\text{RG}_\text{ER}$ model are prefered to $\text{fact}_\text{ENTNLI}$. We notice that scores can be improved and the reports of our model are not systematically better. Following our ideas in Section \ref{sec:limit}, we would like to improve upon our current RRG model in the future. 

\section{Model}\label{app:model}

Encoder: 

\begin{lstlisting}[
    basicstyle=\small,
]
{
    encoder: CNN
    backbone: densenet121
    output_layer: features (of size 49x1024)
    dropout_out: No dropout
    output_size: 1024
}
\end{lstlisting}

The decoder is based on HuggingFace~\cite{wolf2020transformers}:

\begin{lstlisting}[
    basicstyle=\small,
]
{
    decoder: BertGenerationDecoder
    add_cross_attention: true
    attention_probs_dropout_prob: 0.1
    hidden_act: gelu
    hidden_dropout_prob: 0.1
    hidden_size: 768
    initializer_range: 0.02
    intermediate_size: 3072
    is_decoder: true
    layer_norm_eps: 1e-05
    max_position_embeddings: 514
    num_attention_heads: 12
    num_hidden_layers: 1
    position_embedding_type: absolute
    type_vocab_size: 1
    vocab_size: 9877 (for mimic-cxr)

}
\end{lstlisting}

The model has 25.8M learnable parameters and fits on a single GPU of 12GB. \\

The training hyper parameters are as such:

\begin{lstlisting}[
    basicstyle=\small,
]
{
  
  batch_size: 128
  optimizer: RAdam
  optim_params:
    lr: 0.0003
    weight_decay: 0.
  lr_decay: ReduceLROnPlateau
  lr_decay_params:
    factor: 0.8
    patience: 1
    min_lr: 0.000001
    threshold_mode: abs
  early_stop: 10
  early_stop_metric: ROUGEL
}
\end{lstlisting}
The plateau is monitored on ROUGEL metric during eval.

\newpage

\section{Results on the impression section} \label{app:impression}

\begin{table}[!h]
\setlength{\tabcolsep}{5pt}
\renewcommand{\arraystretch}{1.4}
	\centering
	\begin{tabular}{lcccccc}
	        \hline
	    \multicolumn{4}{c}{\textbf{MIMIC-CXR}}\\
        \hline
    Model & $\text{F}_1$cXb & $\text{fact}_\text{ENT}$ & $\text{fact}_\text{ENTNLI}$   \\
        \hline
    ours ($\text{RG}_\text{ER}$) & 54.2 & 33.3 & 30.9 \\
        \hline
    & $\text{RG}_\text{E}$ &  $\text{RG}_\text{ER}$ &  $\text{RG}_{\overline{{\text{ER}}}}$   \\
        \hline
            ours ($\text{RG}_\text{ER}$) & 30.3 & 27.7 & 20.9\\
	\end{tabular}
\caption{Results of our $\text{RG}_\text{ER}$ model trained on generating the \textit{Impression} section of MIMIC-CXR instead of \textit{Findings}.}
\label{table:impression_scores}
\end{table}
\end{document}